%% file: main.tex
\documentclass{article}
\usepackage{iclr2017_conference,times}
\usepackage{url}
\usepackage{graphicx} 

\usepackage{subcaption} 
\usepackage{amsmath}
\usepackage{amssymb}
\usepackage{graphicx}
\usepackage[
singlelinecheck=false 
]{caption}

\usepackage{natbib}

\usepackage{algorithm}
\usepackage{algorithmic}

\usepackage[titletoc,toc,title]{appendix}

\usepackage[justification=centering]{caption}

\newcommand{\yrcite}[1]{\citeyearpar{#1}}

\def\abovestrut#1{\rule[0in]{0in}{#1}\ignorespaces}
\def\belowstrut#1{\rule[-#1]{0in}{#1}\ignorespaces}

\def\abovespace{\abovestrut{0.20in}}

\def\belowspace{\belowstrut{0.10in}}

\newcommand{\removed}[1]{}

\usepackage{soul}
\newcommand{\Note}[1]{}
\definecolor{brilliantlavender}{rgb}{0.96, 0.73, 1.0}
\definecolor{thistle}{rgb}{0.847, 0.749, 0.847}
\newcommand{\NoteSigned}[3]{{\sethlcolor{#2}\Note{#1: #3}}}
\newcommand{\NoteHK}[1]{\NoteSigned{HK}{brilliantlavender}{#1}}

\DeclareMathOperator*{\minimize}{minimize}
\DeclareMathOperator{\tr}{Tr}
\DeclareMathOperator*{\argmax}{argmax}

\title{Deep Generalized Canonical Correlation Analysis}

\author{Adrian Benton, Huda Khayrallah, Biman Gujral,  \\
\textbf{Dee Ann Reisinger, Sheng Zhang, Raman Arora} \\
Center for Language and Speech Processing \\
Johns Hopkins University\\
Baltimore, MD 21218, USA \\
\texttt{adrian$^{\dagger}$,huda$^{\star}$,bgujral1$^{\star}$,reisinger$^{\diamond}$,zsheng2$^{\star}$,arora$^{\dagger}$} \\
$^{\star}$\texttt{@jhu.edu},\enspace $^{\diamond}$\texttt{@cogsci.jhu.edu},\enspace $^{\dagger}$\texttt{@cs.jhu.edu}
}

\begin{document}

\maketitle

\begin{abstract} 
We present Deep Generalized Canonical Correlation Analysis (DGCCA) -- a method for learning nonlinear transformations of arbitrarily many views of data, such that the resulting transformations are maximally informative of each other.  While methods for nonlinear two-view representation learning (Deep CCA, \citep{originalDCCA}) and linear many-view representation learning (Generalized CCA~\citep{GCCA1}) exist, DGCCA is the first CCA-style multiview representation learning technique that combines the flexibility of nonlinear (deep) representation learning with the statistical power of incorporating information from many independent sources, or views. 
We present the DGCCA formulation as well as an efficient stochastic optimization algorithm for solving it.  We learn DGCCA representations on two distinct datasets for three downstream tasks: phonetic transcription from acoustic and articulatory measurements, and recommending hashtags and friends on a dataset of Twitter users.  We find that DGCCA representations soundly beat existing methods at phonetic transcription and hashtag recommendation, and in general perform no worse than standard linear many-view techniques.
\end{abstract}

\input{intro_ab}

\section{Prior Work}
\label{Prior Work}

Some of most successful techniques for multiview representation learning are based on canonical correlation analysis~\citep{Wang_15a,Wang_15b} and its extension to the nonlinear and many view settings, which we describe in this section. For other related multiview learning techniques, see Section~\ref{sec:other}.
\subsection{ Canonical correlation analysis (CCA)}
 Canonical correlation analysis (CCA) \citep{cca} is a statistical method that finds maximally correlated linear projections of two random vectors and is a fundamental multiview learning technique.
Given two input views, $X_1 \in \mathbb{R}^{d_1} \text{ and } X_2 \in \mathbb{R}^{d_2}$, with covariance matrices, $\Sigma_{11} \text{ and } \Sigma_{22}$, respectively, and cross-covariance matrix, $\Sigma_{12}$, CCA finds directions that maximize the correlation between them: 
\vspace*{-5pt}
\begin{align*}
\begin{split}
	(u_1^*,u_2^*) &= \argmax_{u_1\in  \mathbb{R}^{d_1}  ,u_2 \in  \mathbb{R}^{d_2}} corr(u_1^\top X_1, u_2^\top X_2)  
    = \argmax_{u_1\in  \mathbb{R}^{d_1}  ,u_2 \in  \mathbb{R}^{d_2}} \frac{u_1^\top \Sigma_{12}u_2}{\sqrt{u_1^\top \Sigma_{11}u_1 u_2^\top\Sigma_{22}u_2}}
\end{split}
\end{align*}
Since this formulation is invariant to affine transformations of $u_1$ and $u_2$, we can write it as the following constrained optimization formulation: 
\begin{align}
 \label{eq:cca}
 \begin{split}
 (u_1^*,u_2^*) &= \argmax_{u_1^\top\Sigma_{11}u_1=u_2^\top\Sigma_{22}u_2=1} u_1^\top\Sigma_{12}u_2 
\end{split}
\end{align}
 This technique has two limitations
 that have led to significant extensions: First, it is limited to learning
 representations that are \emph{linear} transformations of the data in each
 view, and second, it can only leverage two input views. 
 \subsection{Deep Canonical correlation analysis (DCCA)}
 Deep CCA (DCCA) \citep{originalDCCA} is an extension of CCA that
 addresses the first limitation by finding
 maximally linearly  correlated  \textit{ non-linear transformations }  of two vectors.
 It does this by passing each of the input views through stacked non-linear representations
 and performing CCA on the outputs.
 
Let us use $f_1(X_1)$ and $f_2(X_2)$ to represent the network outputs. The weights, $W_1$ and $W_2$,  of these networks are trained through
 standard backpropagation to maximize the CCA objective:
 
 \begin{align*}
\begin{split}
	(u_1^*,u_2^*,W^*_1, W^*_2) &= \argmax_{u_1,u_2} corr(u_1^\top f_1(X_1), u_2^\top f_2(X_2))  \\ 
\end{split}
\end{align*}
 DCCA is still limited to only 2 input views. 
 
 \subsection{Generalized Canonical correlation analysis (GCCA)}
 \label{gcca}
Another extension of CCA, which addresses the limitation on the number of
 views, is Generalized CCA (GCCA) \citep{GCCA1}. 
 It corresponds to solving the optimization problem in Equation (\ref{eq:gcca}), of finding a shared representation $G$ of $J$ different views, where $N$ is the number of data points, $d_j$ is the dimensionality of the $j$th view, $r$ is the dimensionality of the learned representation, and $X_j \in \mathbb{R}^{d_j \times N}$ is the data matrix for the $j$th view.\footnote{Horst's original presentation of GCCA is in a very different form from the one used here but has been shown to be equivalent \citep{kettenring71}.} 
 \begin{align}
\label{eq:gcca}
\begin{split}
	\minimize_{U_j \in \mathbb{R}^{d_j \times r}, G \in \mathbb{R}^{r
	\times N}} &\sum_{j=1}^J \|G - U_j^\top X_j\|_F^2  \\ \text{subject
	to} \qquad &GG^\top = I_r
\end{split}
\end{align}
Solving GCCA requires finding an eigendecomposition of an $N \times N$ matrix, which scales quadratically with sample size and leads to memory constraints. Unlike CCA and DCCA, which only learn projections or transformations 
on each of the views, GCCA also learns a view-independent representation
$G$ that best reconstructs all of the view-specific representations
simultaneously. The key limitation of GCCA is that it can only learn \emph{linear} transformations of each view.  

\input{dgcca}

\section{Experiments}
\label{sec:results}

\vspace*{-5pt}
\subsection{Synthetic Multiview Mixture Model}
\label{sect:synthdat}
\begin{figure*}[h]
\begin{center}
\includegraphics[height=0.15\textheight]{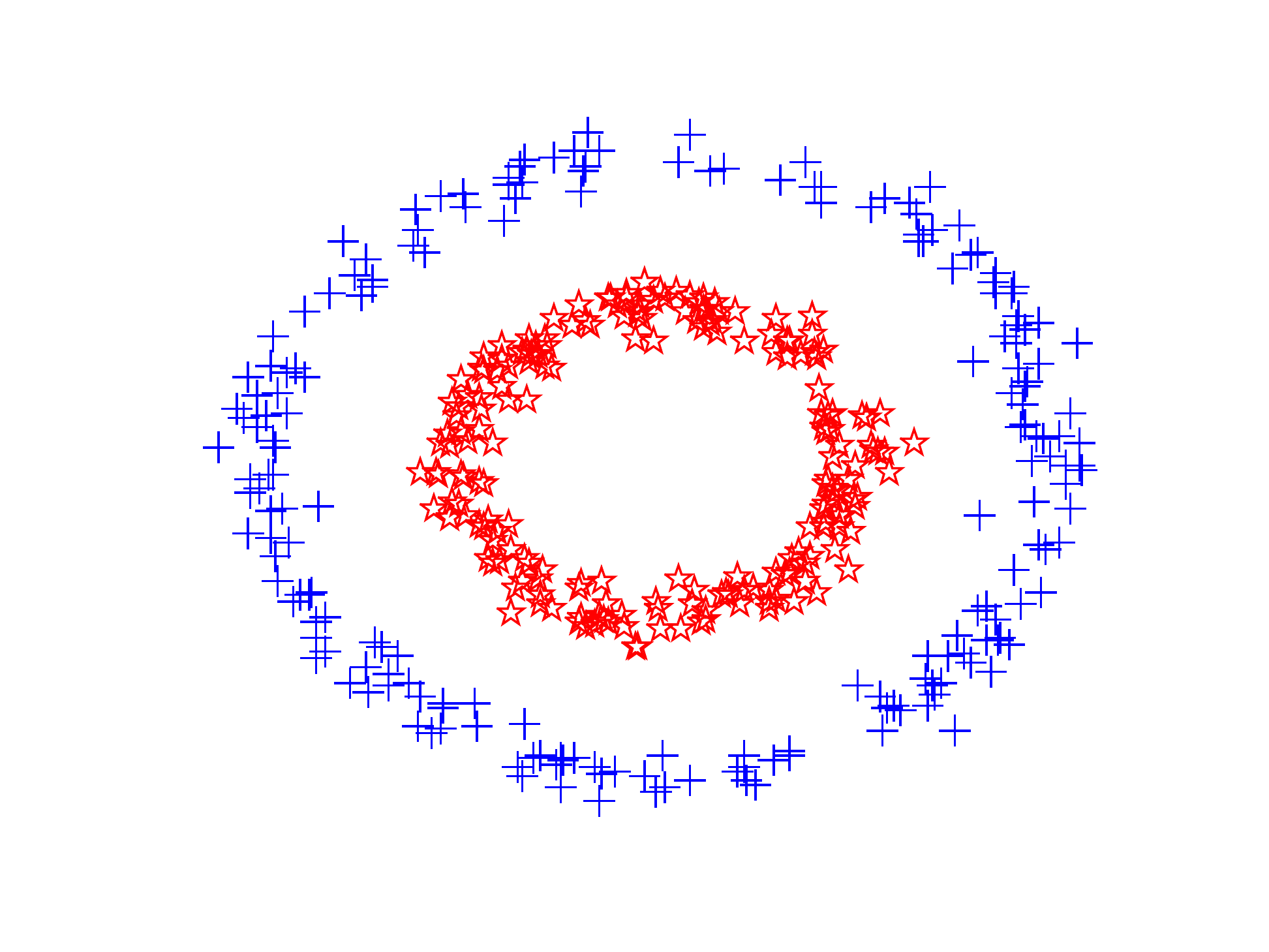}
\includegraphics[height=0.15\textheight]{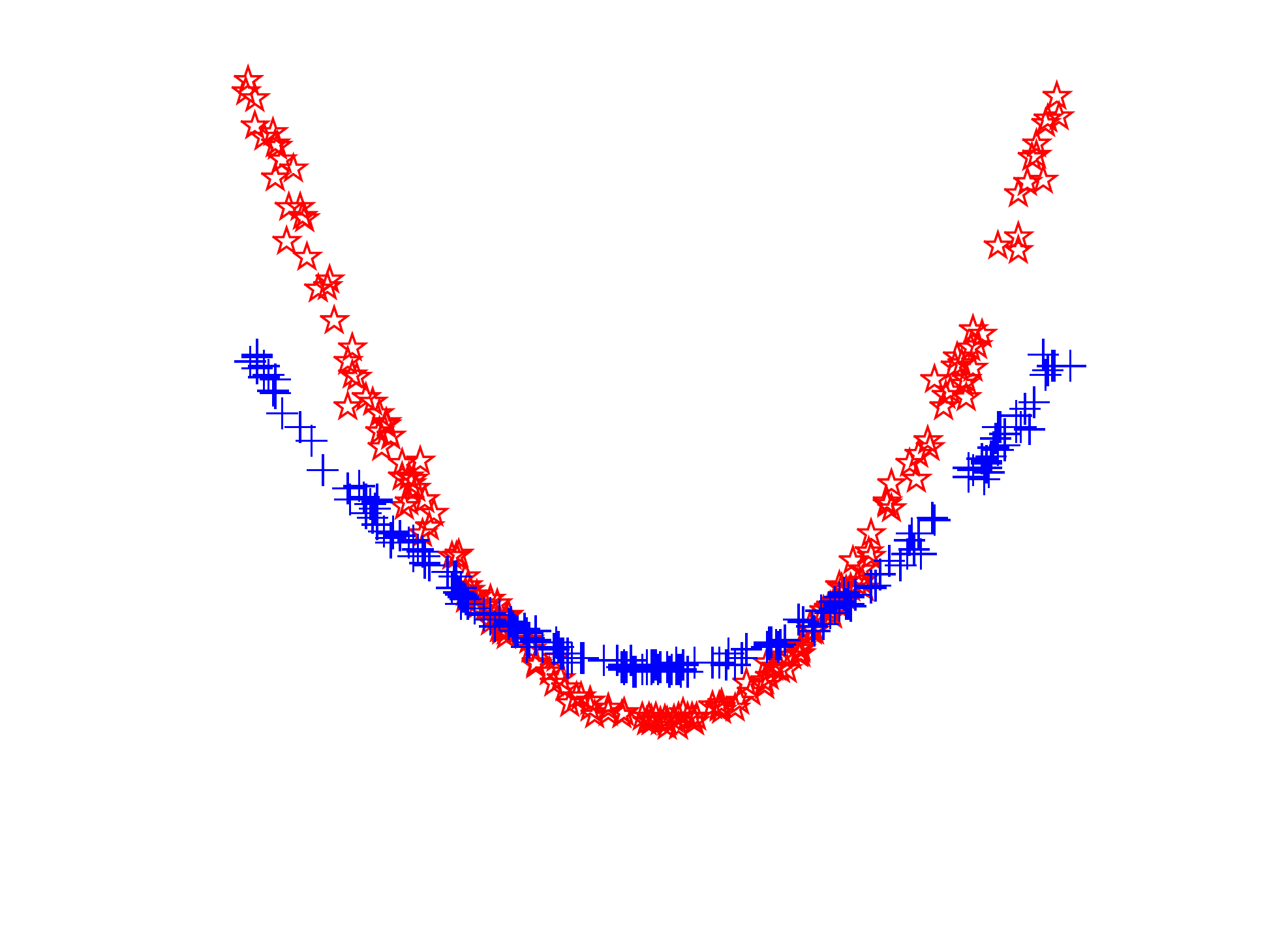}
\includegraphics[height=0.15\textheight]{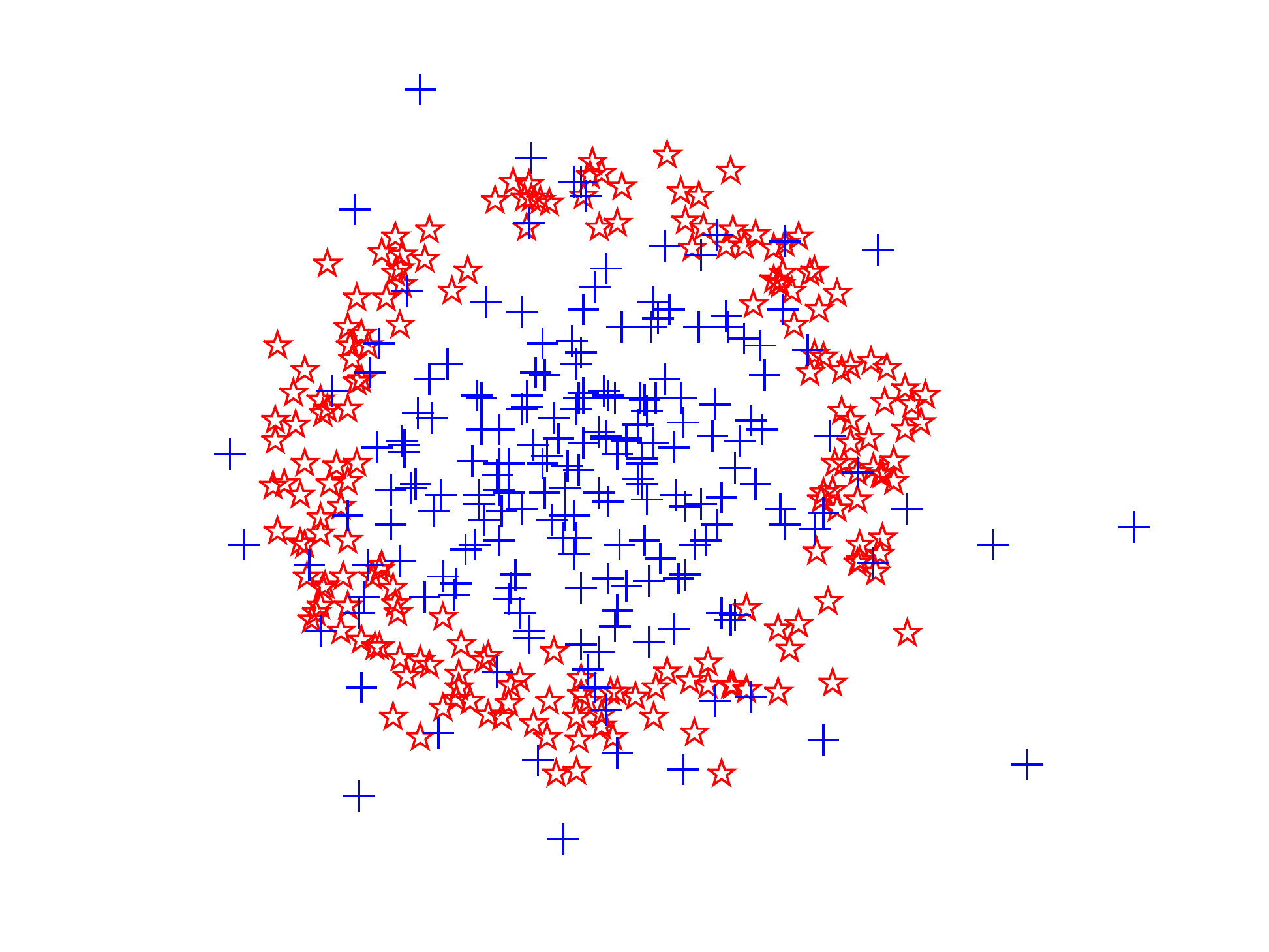}
\vspace{-6mm}
\end{center}
\caption{Synthetic data used in in Section \ref{sect:synthdat} experiments.}
\vspace{-4mm}
\label{fig:synthdat}
 \end{figure*}

In this section, we apply DGCCA to a small synthetic data set to show how it preserves
the generative structure of data sampled from a multiview mixture model. The
data we use for this experiment are plotted in Figure \ref{fig:synthdat}. Points
that share the same color across different views are sampled from the same
mixture component. 

\begin{figure}[h]
\begin{center}
\begin{tabular}{cc}
\includegraphics[height=0.26\textwidth]{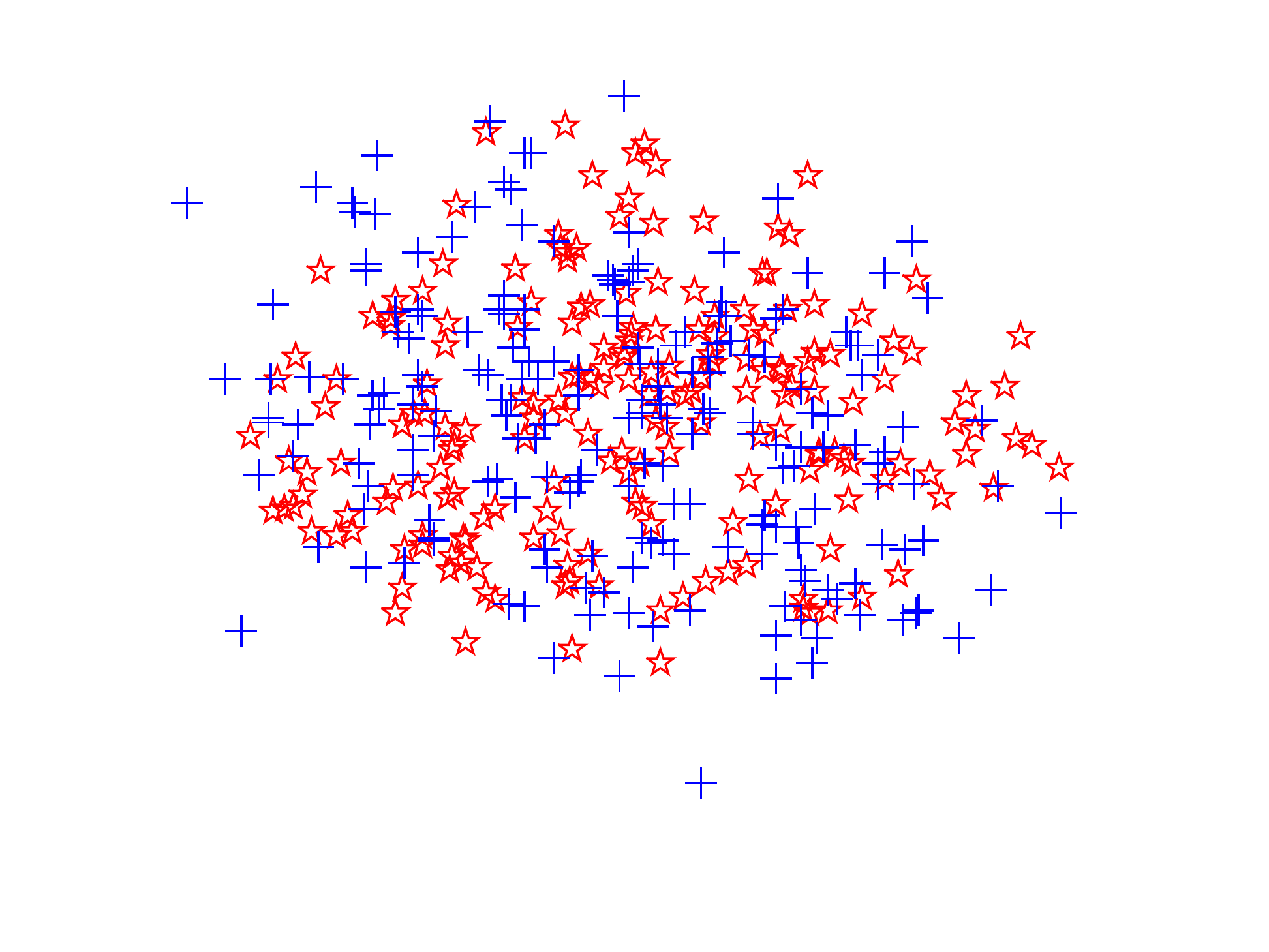} & \hspace*{-20pt} \includegraphics[height=0.26\textwidth]{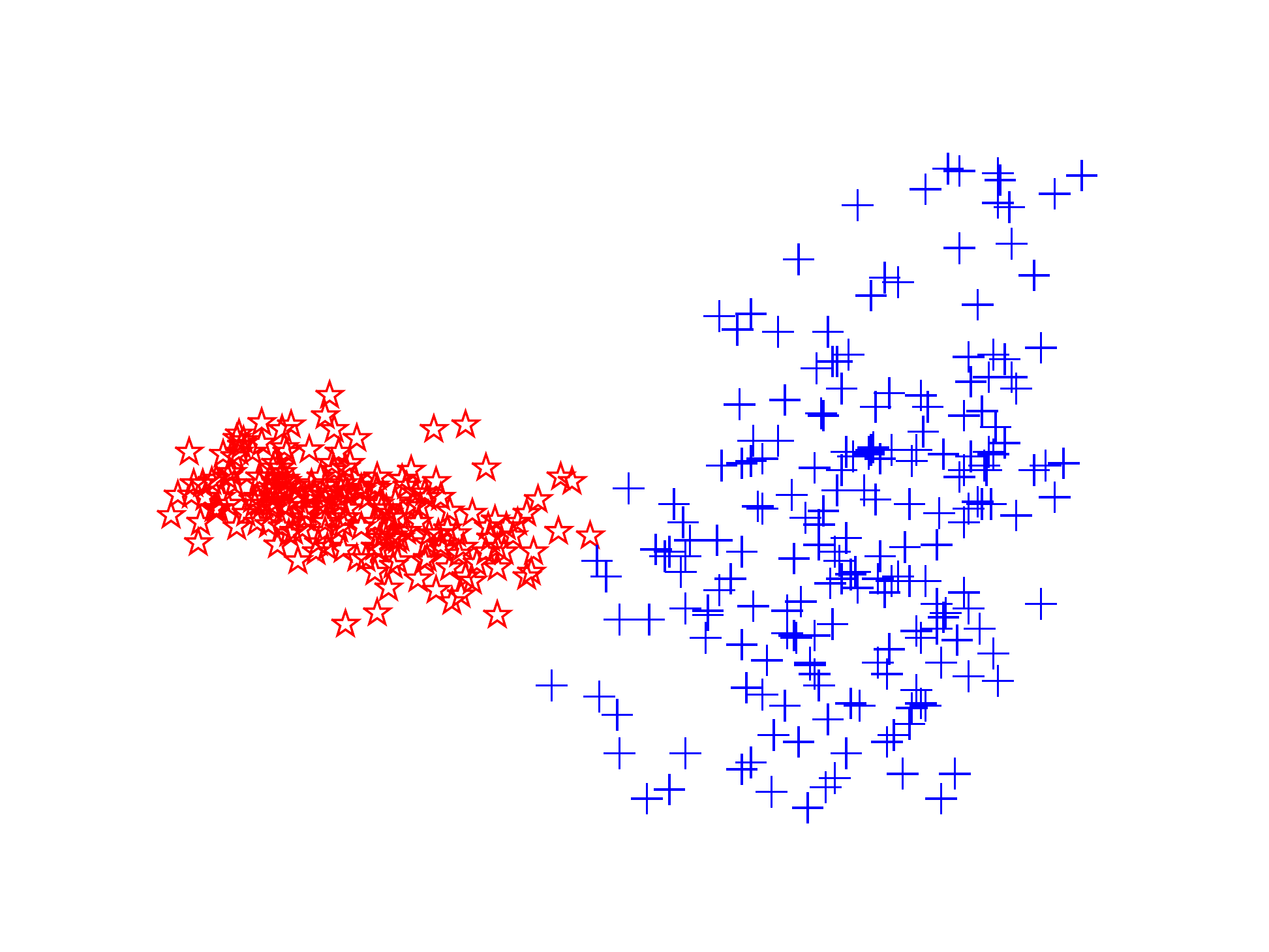}\\
(a) GCCA & (b) DGCCA
\end{tabular}
\end{center}
\vspace{-4mm}
\caption{The matrix $G$ learned from applying (linear) GCCA or DGCCA to the data in Figure \ref{fig:synthdat}.}
\label{fig:synthling}
\vspace{-4mm}
\end{figure}

Importantly, in each view, there is no linear transformation
of the data that separates the two mixture components, in the sense that
the generative structure of the data could not be exploited by a linear model.
This point is reinforced by Figure~\ref{fig:synthling}(a), which shows the
two-dimensional representation $G$ learned by applying (linear) GCCA to
the data in Figure \ref{fig:synthdat}. The learned representation 
completely loses the structure of the data.

We can contrast the failure of GCCA to preserve structure with the result
of applying DGCCA; in this case, the input neural networks had three hidden layers with ten units each with weights randomly initialized.
We plot the representation $G$ learned by DGCCA in Figure
\ref{fig:synthling} (b). In this representation, the mixture components are easily
separated by a linear classifier; in fact, the structure is largely preserved
even after projection onto the first coordinate of G.

\begin{figure*}[h]
\begin{center}
\includegraphics[height=0.15\textheight]{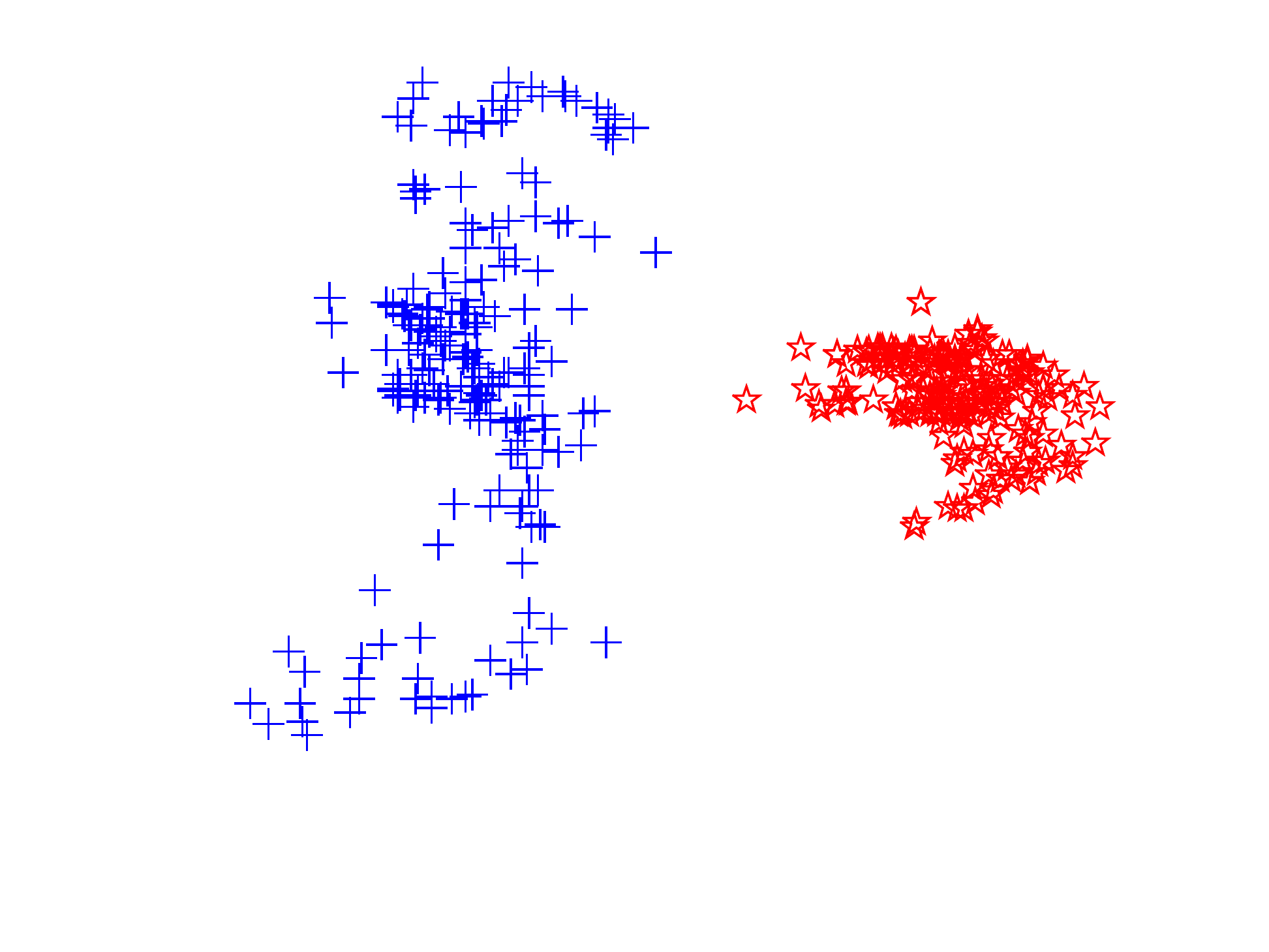}
\includegraphics[height=0.15\textheight]{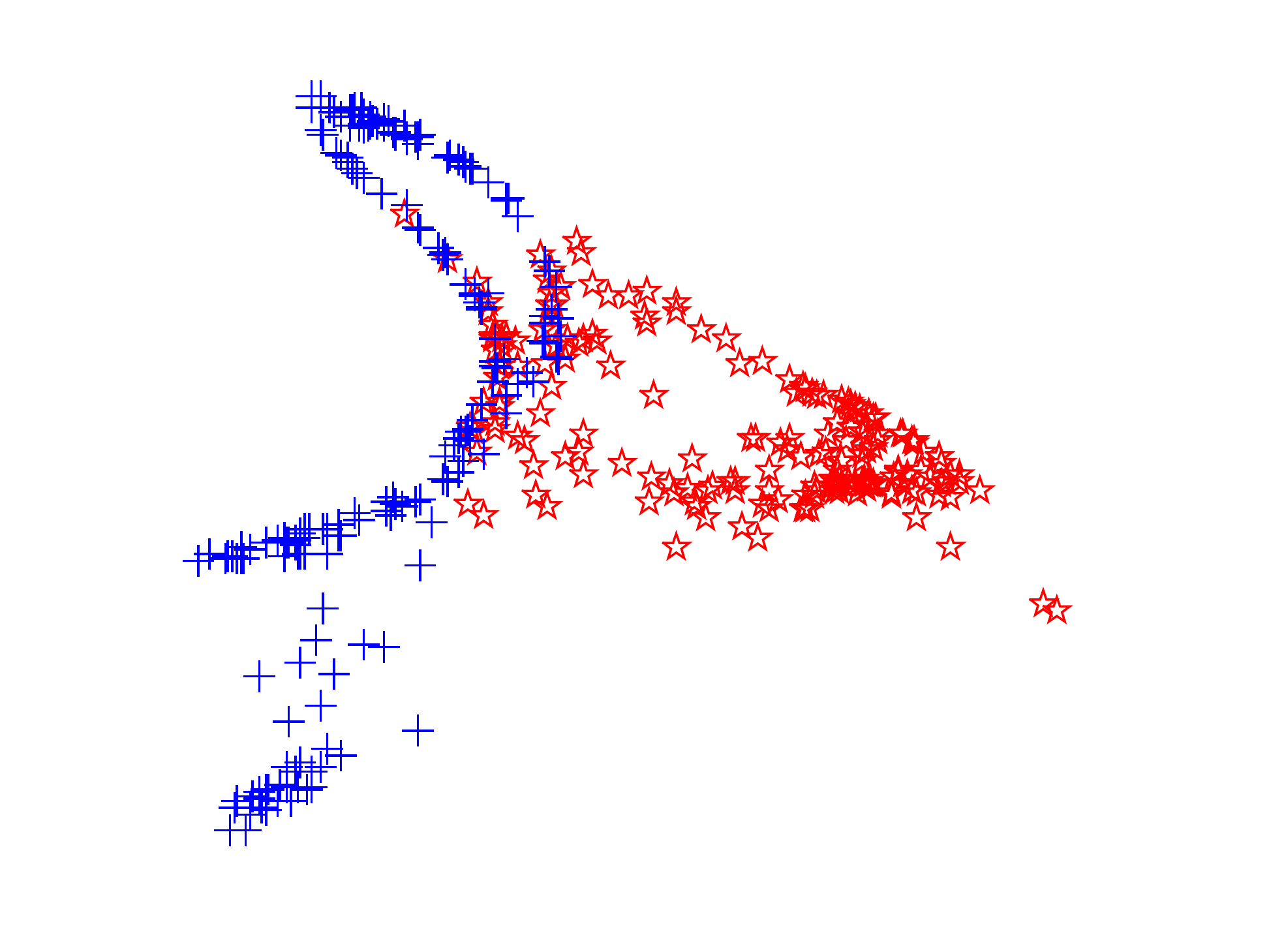}
\includegraphics[height=0.15\textheight]{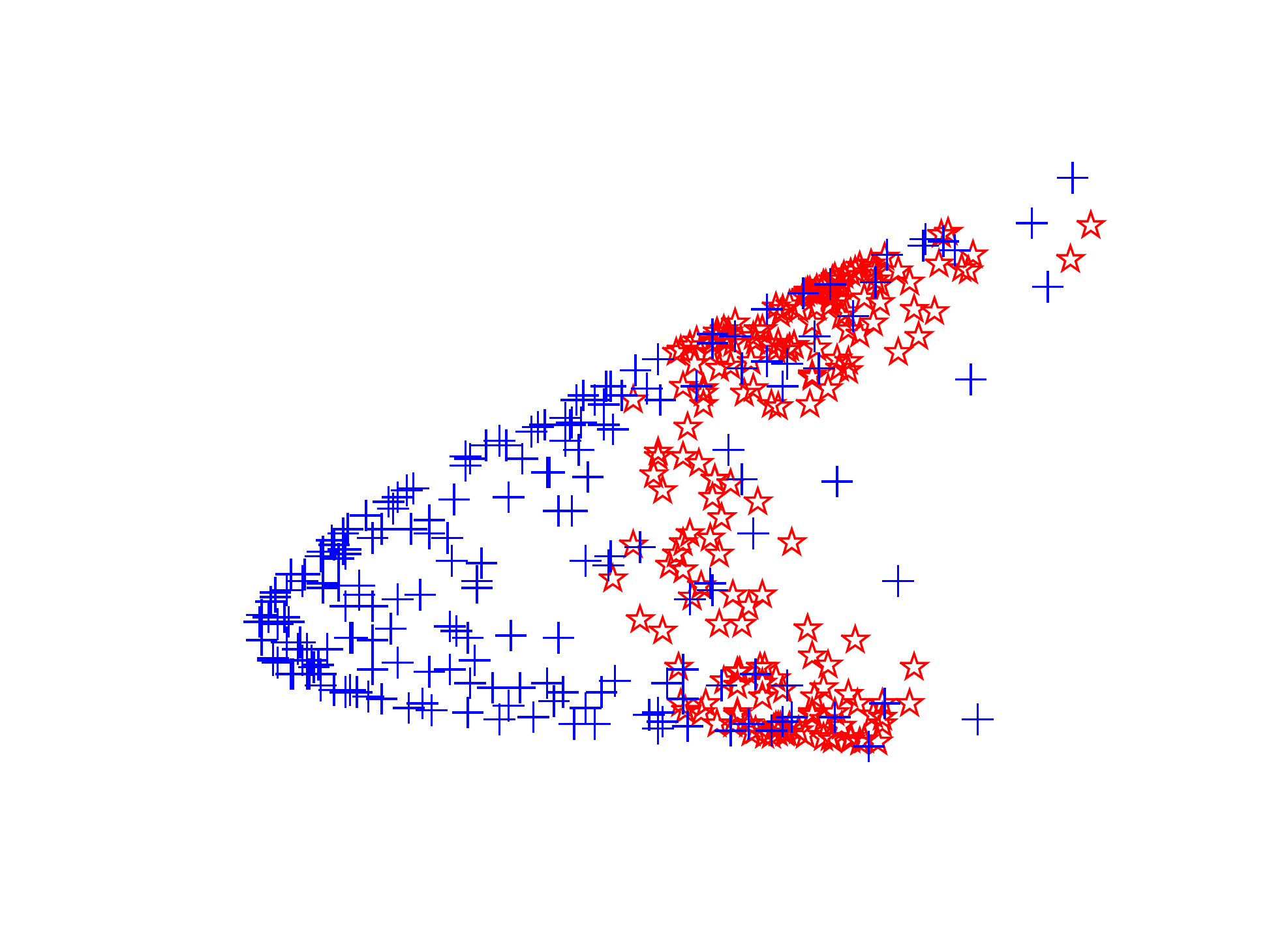}
\vspace{-8mm}
\end{center}
\caption{Outputs of the trained input neural networks in Section \ref{sect:synthdat} applied to the data in Figure \ref{fig:synthdat}.}
\label{fig:synthnn}
\end{figure*}

It is also illustrative to consider the view-specific representations
learned by DGCCA, that is, to consider the outputs of
the neural networks that were trained to maximize the GCCA objective.
We plot the representations in Figure \ref{fig:synthnn}.
For each view, we have learned a nonlinear mapping that does remarkably
well at making the mixture components linearly separable. Recall that absolutely no direct supervision was given about which
mixture component each point was generated from. The only training signals available
to the networks were the reconstruction errors between the network outputs and 
the learned representation $G$.

\vspace{-4mm}

\subsection{Phoneme Classification}
In this section, we discuss experiments on the University of Wisconsin X-ray Microbeam Database (XRMB) \citep{data}. XRMB contains acoustic and articulatory recordings as well as phonemic labels. We present phoneme classification results on the acoustic vectors projected using DCCA, GCCA, and DGCCA. We set acoustic and articulatory data as the two views and phoneme labels as the third view for GCCA and DGCCA. For classification, we run K-nearest neighbor classification \citep{knn} on the projected result. 
\subsubsection{Data}
\vspace*{-5pt}
We use the same train/tune/test split of the data as \citet{labels}. To limit experiment runtime, we use a subset of speakers for our experiments. We run a set of \emph{cross-speaker} experiments using the male speaker JW11 for training and two splits of JW24 for tuning and testing. We also perform parameter tuning for the third view with 5-fold cross validation using a \emph{single speaker}, JW11. For both experiments, we use acoustic and articulatory measurements as the two views in DCCA. Following the pre-processing in \citet{originalDCCA}, we get 273 and 112 dimensional feature vectors for the first and second view respectively. Each speaker has $\sim$50,000 frames. For the third view in GCCA and DGCCA, we use 39-dimensional one-hot vectors corresponding to the labels for each frame, following \citet{labels}.
\vspace*{-5pt}
\subsubsection{Parameters }
\label{Parameters}
\vspace*{-5pt}
We use a fixed network size and regularization for the first two views, each containing three hidden layers with sigmoid activation functions.  Hidden layers for the acoustic view were all width 1024, and layers in the articulatory view all had width 512 units. L2 penalty constants of 0.0001 and 0.01 were used to train the acoustic and articulatory view networks, respectively. The output layer dimension of each network is set to 30 for DCCA and DGCCA. For the 5-fold speaker-dependent experiments, we performed a grid search for the network sizes in \{$128,256,512,1024$\} and covariance matrix regularization in \{$10^{-2},10^{-4},10^{-6},10^{-8}$\} for the third view in each fold.  We fix the hyperparameters for these experiments optimizing the networks with minibatch stochastic gradient descent with a step size of 0.005, batch size of 2000, and no learning decay or momentum. The third view neural network had an L2 penalty of 0.0005.
\vspace*{-5pt}
\subsubsection{Results}
\vspace*{-5pt}
As we show in Table \ref{tbl:phone_perf}, DGCCA improves upon both the linear multiview GCCA and the non-linear 2-view DCCA for both the cross-speaker and speaker-dependent cross-validated tasks.

In addition to accuracy, we examine the reconstruction error, i.e. the objective in Equation~\ref{eq:dgcca}, obtained from the objective in GCCA and DGCCA.\footnote{For 2-view experiments, correlation is a common metric to compare performance. Since that metric is unavailable in a multiview setting, reconstruction error is the analogue.} This sharp improvement in reconstruction error shows that a non-linear algorithm can better model the data. 

In this experimental setup, DCCA under-performs the baseline of simply running KNN on the original acoustic view. Prior work considered the output of DCCA stacked on to the central frame of the original acoustic view (39 dimensions). This poor performance, in the absence of original features, indicates that it was not able to find a more  informative projection than original acoustic features based on correlation with the articulatory view within the first 30 dimensions.

\begin{table}[h]
\centering
\caption{KNN phoneme classification performance}
\label{tbl:phone_perf}
\begin{small}
\begin{sc}
\begin{tabular}{lcccccc}
\hline
\abovespace\belowspace
		 & \multicolumn{3}{c}{Cross-Speaker} & \multicolumn{3}{c}{Speaker-dependent} \\
		 & Dev  & Test  & Rec & Dev  & Test  & Rec \\
Method &  Acc &  Acc &  Error &  Acc &  Acc &  Error \\
\hline
\abovespace
MFCC  &  48.89 & 49.28 & & 66.27 & 66.22 & \\
DCCA  &  45.40 & 46.06 & & 65.88 & 65.81 & \\
GCCA  &  49.59 & 50.18 & 40.67 & 69.52 & 69.78 & 40.39 \\
DGCCA &  \textbf{53.78} & \textbf{54.22} & \textbf{35.89} & \textbf{72.62} & \textbf{72.33} & \textbf{20.52} \\
\hline
\end{tabular}
\end{sc}
\end{small}
\vskip -0.1in

\end{table}

 \begin{figure*}[t]
\begin{center}
\centering
\begin{tabular}{cc}
\includegraphics[height=0.35\textwidth]{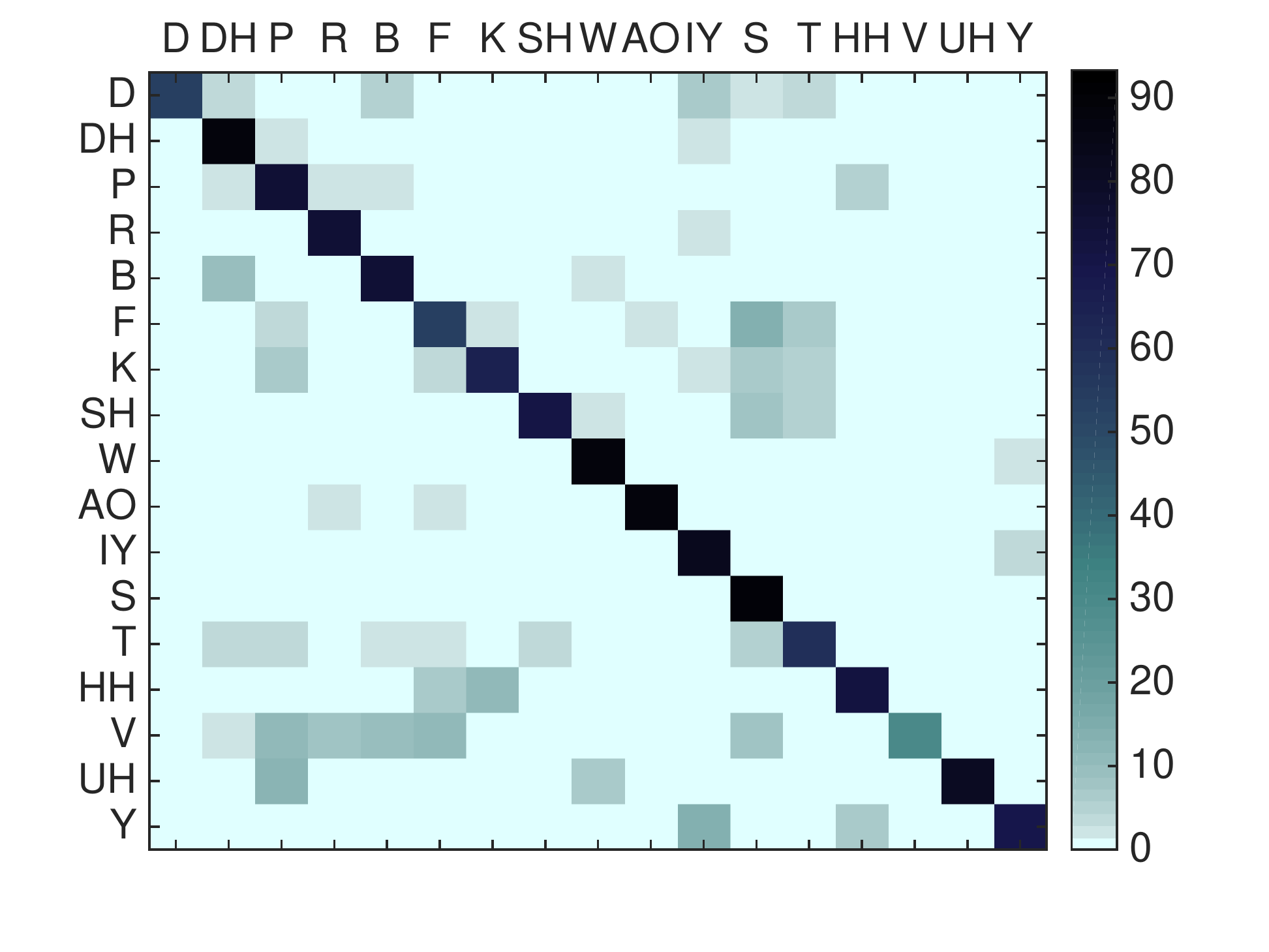} & \hspace*{-20pt} \includegraphics[height=0.35\textwidth]{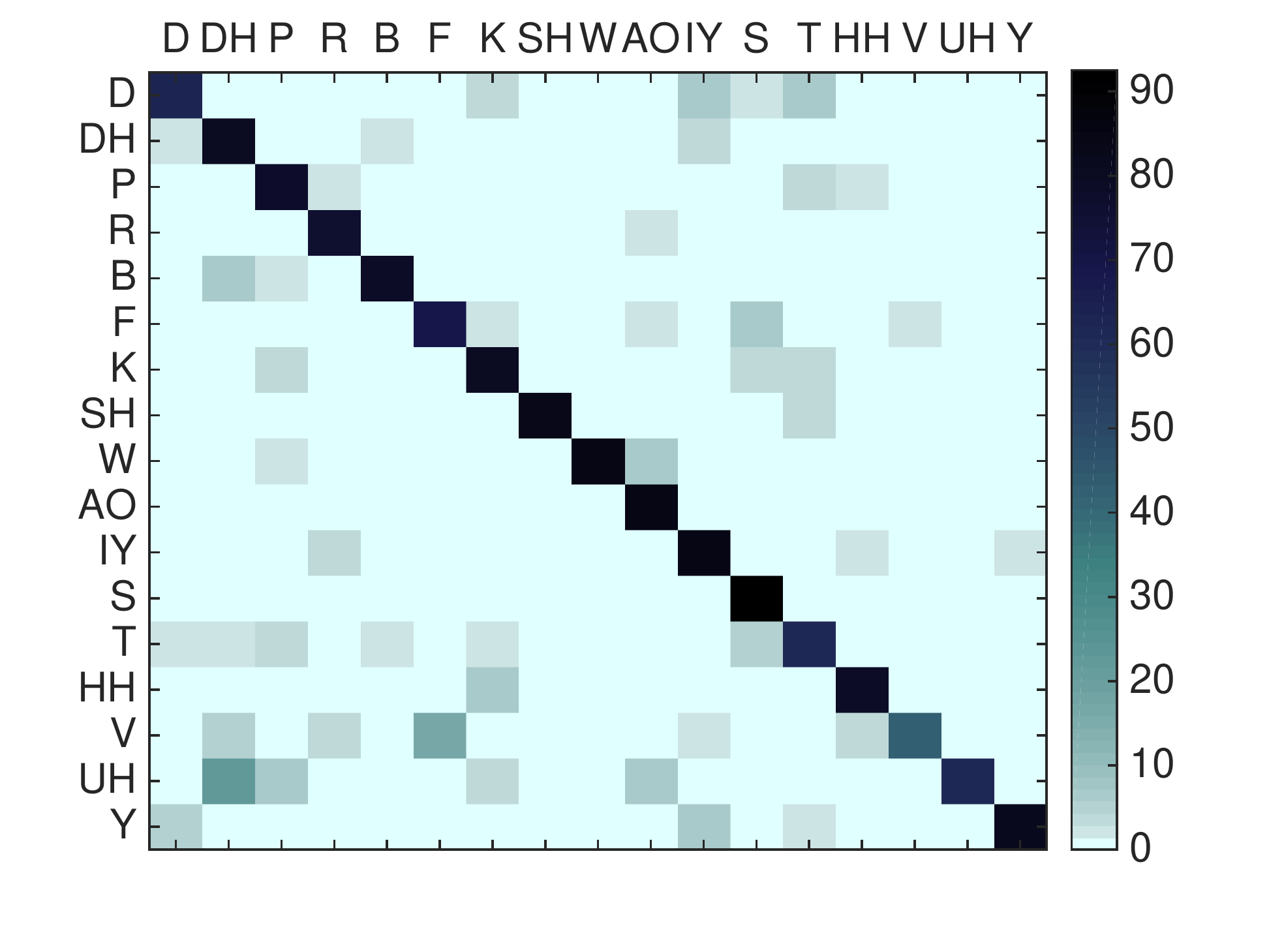}\\
(a) GCCA & (b) DGCCA
\end{tabular}
\end{center}
\vspace{-4mm}
\caption{The confusion matrix for speaker-dependent GCCA and DGCCA}
\label{fig:CMAT_GCCA}
\vspace{-5mm}
\end{figure*}

To highlight the improvements of DGCCA over GCCA, Figure \ref{fig:CMAT_GCCA} presents a subset of the the confusion matrices on speaker-dependent test data. In particular, we observe large improvements in the classification of $D$, $F$, $K$, $SH$, $V$ and $Y$. GCCA outperforms DGCCA for $UH$ and $DH$. These matrices also highlight the common misclassifications that DGCCA improves upon. For instance, DGCCA rectifies the frequent misclassification of $V$ as $P$, $R$ and $B$ by GCCA. In addition, commonly incorrect classification of phonemes such as $S$ and $T$ is corrected by DGCCA, which enables better performance on other voiceless consonants such as like $F$, $K$ and $SH$. Vowels are classified with almost equal accuracy by both the methods.

\vspace*{-5pt}
\subsection{Twitter User Hashtag \& Friend Recommendation}
\label{sec:results_twitter}
\vspace*{-5pt}

Linear multiview techniques are effective at recommending hashtag and friends for
Twitter users \citep{tusersmv}.  In this experiment, six views of a Twitter user were
constructed by applying principal component analysis (PCA) to the bag-of-words
representations of (1) tweets posted by the ego user, (2) other mentioned users, (3) their
friends, and (4) their followers, as well as one-hot encodings of the local (5) friend and
(6) follower networks.  We learn and evaluate DGCCA models on identical training,
development, and test sets as \citet{tusersmv}, and evaluate the DGCCA
representations on macro precision at 1000 (P@1000) and recall at 1000 (R@1000) for the
hashtag and friend recommendation tasks described there.

We trained 40 different DGCCA model architectures, each with identical architectures across views, where the width of the hidden and output layers, $c_{1}$ and $c_{2}$, for each view
are drawn uniformly from $[10,1000]$, and the auxiliary representation width $r$ is drawn
uniformly from $[10,c_{2}]$\footnote{We chose to restrict ourselves to a single hidden layer with
non-linear activation and identical architectures for each view, so as to avoid a fishing
expedition.  If DGCCA is appropriate for learning
Twitter user representations, then a good architecture should require little exploration.}.
All networks used ReLUs as activation functions, and were
optimized with Adam \citep{kingma2014adam} for 200 epochs\footnote{From preliminary
experiments, we found that Adam pushed down reconstruction
error more quickly than SGD with momentum, and that ReLUs were easier to optimize than
sigmoid activations.}.
Networks were trained on 90\% of 102,328 Twitter users, with 10\% of users used as a
tuning set to estimate heldout reconstruction error for model selection.
We report development and test results for the best performing model on the downstream task
development set.  Learning rate was set to $10^{-4}$ with an L1 and L2 regularization constants of $0.01$ and $0.001$ for all weights \footnote{This setting of regularization constants led to low reconstruction error in preliminary experiments.}.

\begin{table}[h]
\vspace{-2mm}
\vspace*{-1pt}
\caption{Dev/test performance at Twitter friend and hashtag recommendation tasks.}
\label{tbl:twitter_perf}
\vspace*{-5pt}
\vspace{-2mm}
\begin{center}
\begin{small}
\begin{sc}
\begin{tabular}{lcccc}
\hline
\abovespace
& \multicolumn{2}{c}{Friend} & \multicolumn{2}{c}{Hashtag} \\
Algorithm & P@1000 & R@1000 & P@1000 & R@1000 \\
\hline
\abovespace
PCA[text+net] & \textbf{0.445/0.439} & \textbf{0.149/0.147} & 0.011/0.008 & 0.312/0.290 \\
GCCA[text] & 0.244/0.249 & 0.080/0.081 & 0.012/0.009 & 0.351/0.326 \\
GCCA[text+net] & 0.271/0.276 & 0.088/0.089 & \textbf{0.012/0.010} & 0.359/0.334 \\
DGCCA[text+net] & 0.297/0.268 & 0.099/0.090 & \textbf{0.013/0.010} & \textbf{0.385/0.373} \\
\hline
\abovespace
WGCCA[text] & 0.269/0.279 & 0.089/0.091 & 0.012/0.009 & 0.357/0.325 \\
WGCCA[text+net] & 0.376/0.364 & 0.123/0.120 & 0.013/0.009 & 0.360/0.346 \\
\hline
\end{tabular}
\end{sc}
\end{small}
\end{center}
\vspace{-2mm}
\end{table}

Table \ref{tbl:twitter_perf} displays the performance of DGCCA compared to PCA[text+net] (PCA
applied to concatenation of view feature vectors), linear GCCA applied to the four text
views, \emph{[text]}, and all views, \emph{[text+net]}, along with a
weighted GCCA variant (WGCCA).  We learned PCA, GCCA, and WGCCA representations of width
$r \in \{10, 20, 50, 100, 200, 300, 400, 500, 750, 1000\}$, and report the best performing
representations on the development set.

There are several points to note: First is that DGCCA outperforms linear methods at
hashtag recommendation by a wide margin in terms of recall.  This is exciting because this
task was shown to benefit from incorporating more than just two views from Twitter
users.  These results suggest that a nonlinear transformation of the input views can yield
additional gains in performance.  In addition, WGCCA models sweep over every
possible weighting of views with weights in $\{0, 0.25, 1.0\}$.
WGCCA has a distinct advantage in that the model is allowed to discriminatively
weight views to maximize downstream performance.  The fact that
DGCCA is able to outperform WGCCA at hashtag recommendation is encouraging, since WGCCA has
much more freedom to discard uninformative views, whereas the DGCCA objective forces
networks to minimize reconstruction error equally across all views.
As noted in \citet{tusersmv}, only the friend network view was useful for learning
representations for friend recommendation (corroborated by performance of PCA applied to
friend network view), so it is unsurprising that DGCCA when applied to all views cannot
compete with WGCCA representations learned on the single useful friend network view\footnote{The performance of WGCCA suffers compared to PCA because whitening the friend network data ignores the fact that the spectrum of the decays quickly with a long tail -- the first few principal components made up a large portion of the variance in the data, but it was also important to compare users based on other components.}.

\vspace{-2mm}
\vspace*{-8pt}
\section{Other Multiview Learning Work}
\label{sec:other}
\vspace*{-8pt}
There has been strong work outside of CCA-related methods to combine nonlinear representation and learning from multiple views.
\citet{kumar2011co} elegantly outlines two main approaches these methods take to learn a joint representation from many views: either by 1) explicitly
maximizing pairwise similarity/correlation between views or by 2) alternately optimizing a
shared, ``consensus'' representation and view-specific transformations to maximize similarity.  Models such as the siamese network proposed by \citet{masci2014multimodal}, fall in the former camp, minimizing the squared error between embeddings learned from each view, leading to a quadratic increase in the terms of the loss function size as the number of views increase.  \citet{rajendran2015bridge} extend Correlational Neural Networks \citep{chandar2015correlational} to many views and avoid this quadratic explosion in the loss function by only computing correlation between each view embedding and the embedding of a ``pivot'' view.  Although this model may be appropriate for tasks such as multilingual image captioning, there are many datasets where there is no clear method of choosing a pivot view.  The DGCCA objective does not suffer from this quadratic increase w.r.t. the number of views, nor does it require a privileged pivot view, since the shared representation is learned from the per-view representations.

Approaches that estimate a ``consensus'' representation, such as the multiview spectral clustering approach in \citet{kumar2011co}, typically do so by an alternating optimization scheme which depends on a strong initialization to avoid bad local optima.  The GCCA objective our work builds on is particularly attractive, since it admits a globally optimal solution for both the view-specific projections $U_1 \ldots U_J$, and the shared representation $G$ by singular value decomposition of a single matrix: a sum of the per-view projection matrices.  Local optima arise in the DGCCA objective only because we are also learning nonlinear transformations of the input views.  Nonlinear multiview methods often avoid learning these nonlinear transformations by assuming that a kernel or graph Laplacian (e.g. in multiview clustering) is given \citep{kumar2011co,xiaowen2014multi,sharma2012generalized}.

\vspace{-2mm}
\vspace*{-5pt}
\section{Conclusion} 
\label{sec:conclusions}
\vspace*{-5pt}
We present DGCCA, a method for non-linear multiview representation learning from an arbitrary number of views. We show that DGCCA clearly outperforms prior work when using labels as a third view
\citep{originalDCCA,labels, sgd}, and can successfully exploit multiple views to learn
user representations useful for downstream tasks such as hashtag recommendation for Twitter
users.
To date, CCA-style
multiview learning techniques were either restricted to learning representations from no more
than two views, or strictly linear transformations of the input views. This work overcomes
these limitations. 

\bibliography{main}
\bibliographystyle{iclr2017_conference}
\newpage

\begin{appendices}

\section{Deriving the GCCA Objective Gradient}
\label{sect:deriv}
In order to train the neural networks in DGCCA, 
we need to compute the gradient of the GCCA
objective with respect to any one of its input views. This
gradient can then be backpropagated through the input networks
to derive updates for the network weights.

Let $N$ be the number of data points and $J$ the number of views. Let $Y_j \in \mathbb{R}^{c^j_K \times N}$ 
be the data matrix representing the output of the $j$th neural network, i.e. $Y_j=f_j(X_j)$, where $c^j_K$ is the number of neurons in the output layer of the $j$th network. Then, GCCA can be written as the following optimization problem,
where $r$ is the dimensionality of the learned auxiliary representation:
\begin{align*}
	\minimize_{U_j \in \mathbb{R}^{c^j_K \times r}, G \in \mathbb{R}^{r
	\times N}} &\sum_{j=1}^J \|G - U_j^\top Y_j\|_F^2  \\ \text{subject
	to} \qquad &GG^\top = I_r
\end{align*}
It can be shown that the solution is found by solving a certain
eigenvalue problem. In particular, define $C_{jj} =Y_jY_j^\top \in
\mathbb{R}^{c^j_K \times c^j_K}$, $P_j = Y_j^\top C_{jj}^{-1} Y_j$
(note that $P_j$ is symmetric and idempotent), and
$M = \sum_{j=1}^J P_j$ (since each $P_j$ is psd, so is $M$).
Then the rows of $G$ are the top $r$ (orthonormal) eigenvectors of $M$,
and $U_j = C_{jj}^{-1}Y_jG^\top$. Thus, at the minima of the objective,
we can rewrite the reconstruction error as follows:
\begin{align*}
\sum_{j=1}^J \|G - U_j^\top Y_j\|_F^2 &= 
\sum_{j=1}^J \|G - GY_j^\top C_{jj}^{-1}Y_j\|_F^2 \\
&= \sum_{j=1}^J \|G(I_N - P_j)\|_F^2 \\
&= \sum_{j=1}^J \tr[G(I_N - P_j)G^\top] \\
&= \sum_{j=1}^J \tr(I_r) - \tr(GMG^\top) \\
&= Jr - \tr(GMG^\top)
\end{align*}
Note that we can write the rank-1 decomposition of $M$ as
$\sum_{k=1}^N \lambda_k g_kg_k^\top$. Furthermore, since the
$k$th row of $G$ is $g_k$, 
and since the matrix product $Gg_k = \hat{e}_k$,
\[
GMG^\top = \sum_{k=1}^N \lambda_k Gg_k(Gg_k)^\top
= \sum_{k=1}^r \lambda_k \hat{e}_k\hat{e}_k^\top
\]
But this is just an $N \times N$ diagonal matrix containing the
top $r$ eigenvalues of $M$, so we can write the GCCA objective as
\[
Jr - \sum_{i=1}^r \lambda_i(M)
\]
Thus, minimizing the GCCA objective (w.r.t. the weights of the
neural nets) means maximizing the sum of eigenvalues
$\sum_{i=1}^r \lambda_i(M)$, which we will henceforth denote
by $L$.

Now, we will derive an expression for $\frac{\partial L}{\partial Y_j}$ for any view $Y_j$. First, by the chain rule, and using the fact
that $\frac{\partial L}{\partial M} = G^\top G$ \citep{matcookbook},
\begin{align*}
\frac{\partial L}{\partial (Y_j)_{ab}} &=
\sum_{c,d=1}^N \frac{\partial L}{\partial M_{cd}}
\frac{\partial M_{cd}}{\partial(Y_j)_{ab}} \\
&= \sum_{c,d=1}^N (G^\top G)_{cd}
\frac{\partial M_{cd}}{\partial(Y_j)_{ab}}
\end{align*}
Since $M = \sum_{j'=1}^J P_{j'}$, and since the only projection matrix
that depends on $Y_j$ is $P_j$, $\frac{\partial M}{\partial Y_j} =
\frac{\partial P_j}{\partial Y_j}$. Since $P_j = Y_j^\top C_{jj}^{-1}
Y_j$,
\[
(P_j)_{cd} = \sum_{k,\ell=1}^{c^j_K} (Y_j)_{kc}(C_{jj}^{-1})_{k\ell}
(Y_j)_{\ell d}
\]
Thus, by the product rule,
\begin{align*}
\frac{\partial(P_j)_{cd}}{\partial(Y_j)_{ab}} &= \delta_{cb}\sum_{\ell=1}^{c^j_K} (Y_j)_{\ell d}(C_{jj}^{-1})_{a\ell}
~+ \\
&\quad~\delta_{db}\sum_{k=1}^{c^j_K} (Y_j)_{kc}(C_{jj}^{-1})_{ka}
~+ \\
&\quad~\sum_{k,\ell=1}^{c^j_K} (Y_j)_{kc}(Y_j)_{\ell d}
\frac{\partial(C_{jj}^{-1})_{k\ell}}{\partial(Y_j)_{ab}} \\
&= \delta_{cb}(C_{jj}^{-1}Y_j)_{ad} + \delta_{db}(C_{jj}^{-1}Y_j)_{ac}\\
&\quad~+~\sum_{k,\ell=1}^{c^j_K} (Y_j)_{kc}(Y_j)_{\ell d}
\frac{\partial(C_{jj}^{-1})_{k\ell}}{\partial(Y_j)_{ab}}
\end{align*}
The derivative in the last term can also be computed using the chain
rule:
\begin{align*}
\frac{\partial(C_{jj}^{-1})_{k\ell}}{\partial(Y_j)_{ab}} &=
\sum_{m,n = 1}^{N} \frac{\partial(C_{jj}^{-1})_{k\ell}}
{\partial(C_{jj})_{mn}} \frac{\partial(C_{jj})_{mn}}{\partial(Y_j)_{ab}}
\\
&= -\sum_{m,n = 1}^{N} \Big\{(C_{jj}^{-1})_{km}(C_{jj}^{-1})_{n\ell} \\
&\quad~[\delta_{am}(Y_j)_{nb} + \delta_{an}(Y_j)_{mb}]\Big\} \\
&= -\sum_{n=1}^N (C_{jj}^{-1})_{ka}(C_{jj}^{-1})_{n\ell}(Y_j)_{nb}\\
&\quad~-\sum_{m=1}^N (C_{jj}^{-1})_{km}(C_{jj}^{-1})_{a\ell}(Y_j)_{mb}\\
&= -(C_{jj}^{-1})_{ka}(C_{jj}^{-1}Y_j)_{\ell b} \\
&\quad~-(C_{jj}^{-1})_{a\ell}
(C_{jj}^{-1}Y_j)_{kb}
\end{align*}
Substituting this into the expression for 
$\frac{\partial(P_j)_{cd}}{\partial(Y_j)_{ab}}$ and
simplifying matrix products, we find
that
\begin{align*}
\frac{\partial(P_j)_{cd}}{\partial(Y_j)_{ab}} &=
\delta_{cb}(C_{jj}^{-1}Y_j)_{ad} + \delta_{db}(C_{jj}^{-1}Y_j)_{ac}\\
&\quad-(C_{jj}^{-1}Y_j)_{ac}(Y_j^\top C_{jj}^{-1}Y_j)_{bd}\\
&\quad-(C_{jj}^{-1}Y_j)_{ad}(Y_j^\top C_{jj}^{-1}Y_j)_{bc}\\
&= (I_N - P_j)_{cb}(C_{jj}^{-1}Y_j)_{ad}~+\\
&\quad~(I_N - P_j)_{db}(C_{jj}^{-1}Y_j)_{ac}
\end{align*}
Finally, substituting this into our expression for
$\frac{\partial L}{\partial (Y_j)_{ab}}$,
we find that
\begin{align*}
\frac{\partial L}{\partial (Y_j)_{ab}}
&= \sum_{c,d=1}^N (G^\top G)_{cd}(I_N - P_j)_{cb}(C_{jj}^{-1}Y_j)_{ad}\\
&~+ \sum_{c,d=1}^N (G^\top G)_{cd}(I_N - P_j)_{db}(C_{jj}^{-1}Y_j)_{ac}\\
&= 2[C_{jj}^{-1}Y_jG^\top G(I_N - P_j)]_{ab}
\end{align*}
Therefore,
\[
\frac{\partial L}{\partial Y_j}
= 2C_{jj}^{-1}Y_jG^\top G(I_N - P_j)
\]
But recall that $U_j = C_{jj}^{-1}Y_jG^\top$. Using this, the gradient
simplifies as follows:
\[
\frac{\partial L}{\partial Y_j}
= 2U_jG - 2U_jU_j^\top Y_j
\]
Thus, the gradient is the difference between the
$r$-dimensional auxiliary representation $G$ embedded into
the subspace spanned by the columns of $U_j$ (the first term) 
and the projection of
the network outputs in $Y_j=f_j(X_j)$ onto said subspace (the second term).
Intuitively, if the auxiliary representation $G$ is far away from
the view-specific representation $U_j^\top f_j(X_j)$, then the
network weights should receive a large update.
\newpage

\section{DGCCA Optimization Pseudocode}

Algorithm \ref{algorithm} contains the pseudocode for the DGCCA optimization algorithm. In practice we use stocastic optimization with minibatches, following \citet{sgd}. 

\label{Algorithm}


\begin{algorithm}[H] 
   \caption{Deep Generalized CCA}
   \label{algorithm}
\begin{algorithmic}
   \STATE {\bfseries Input:} multiview data: $X_1, X_2,\ldots,X_J$, \\ \hspace*{10mm} number of iterations $T$, learning rate $\eta$
   \STATE {\bfseries Output:} $O_1, O_2,\ldots, O_J$
   \STATE Initialize weights $W_1, W_2, \ldots, W_J$\;
   \FOR{iteration $t = 1, 2, \ldots, T$}
   \FOR{each view $j = 1,2,\ldots,J$}
   \STATE $O_j \leftarrow$ forward pass of $X_j$ with weights $W_j$\;
   \STATE mean-center $O_j$\;
   \ENDFOR
   \STATE $U_1,\ldots,U_J, G \leftarrow$ {\tt gcca}($O_1, \ldots, O_J$)\;
   \FOR{each view $j = 1,2,\ldots,J$}
   \STATE $\partial F/\partial O_j \leftarrow U_jU_j^\top O_j - U_j G$\;
   \STATE $\nabla W_j \leftarrow$ {\tt backprop}($\partial F/\partial O_j, W_j$)\;
   \STATE $W_j \leftarrow W_j - \eta\nabla W_j$\;
   \ENDFOR
   \ENDFOR
   \FOR{each view $j = 1,2,\ldots,J$}
   \STATE $O_j \leftarrow$ forward pass of $X_j$ with weights $W_j$\;
   \STATE mean-center $O_j$\;
   \ENDFOR
   \STATE $U_1,\ldots,U_J, G \leftarrow$ {\tt gcca}($O_1, \ldots, O_J$)\;
   \FOR{each view $j = 1...J$}
   \STATE $O_j \leftarrow U_j^\top O_j$\;
   \ENDFOR
\end{algorithmic}
\end{algorithm}
\newpage

\section{Reconstruction Error and Downstream Performance}

\begin{figure*}[h]
\begin{center}
\includegraphics[height=0.30\textheight]{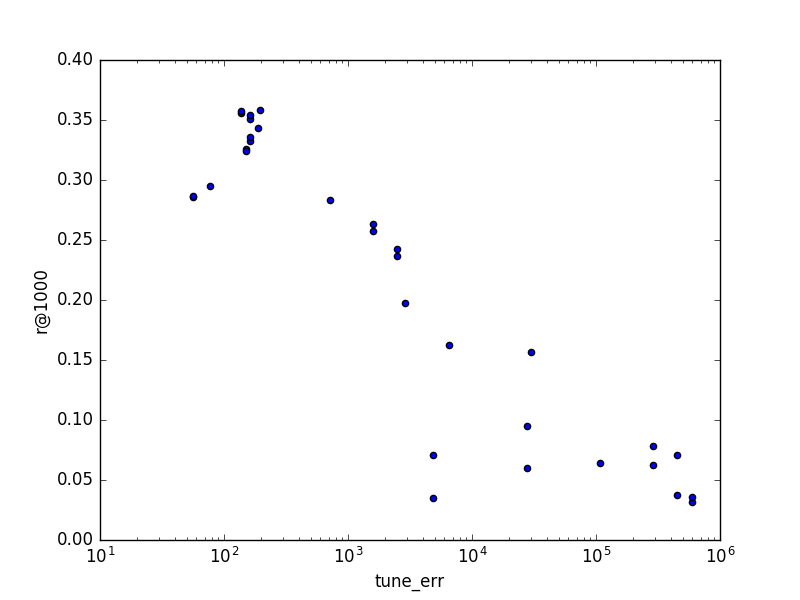}
\end{center}
\caption{Tuning reconstruction error against Recall at 1000 for the hashtag prediction task.  Each point corresponds to a different setting of hyperparameters.}
\label{fig:rec_err}
 \end{figure*}

CCA methods are typically evaluated intrinsically by the amount of correlation captured, or
reconstruction error.  These measures are dependent on the width of the shared embeddings
and view-specific output layers, and do not necessarily predict downstream performance.
Although reconstruction error cannot solely be relied on for model selection for a downstream
task, we found that it was a useful as a signal to weed out very poor models.  Figure
\ref{fig:rec_err} shows the reconstruction error against hashtag prediction Recall at 1000 for
an initial grid search of DGCCA hyperparameters.  Models with tuning reconstruction error
greater than $10^{3}$ can safely be ignored, while there is some variability in the performance
of models with achieving lower error.

Since a DGCCA model with high reconstruction error suggests that the views do not
agree with each other at all, it makes sense that the shared embedding will likely be noisy, whereas a
relatively lowly reconstruction error suggests that the transformed views have converged to a
stable solution.

\end{appendices}

\end{document}

%% file: intro_ab.tex
\section{Introduction}
\label{Introduction}

Multiview representation learning refers to settings where one has access to many ``views''
of data, at train time.  Views often correspond to different modalities or independent information about examples: a scene represented as a series of audio and image frames, a social media user characterized by the messages they post and who they friend, or a speech utterance and the configuration of the speaker's tongue.  Multiview techniques learn a representation of data that captures the sources of variation common to all views.

Multiview representation techniques are attractive for intuitive reasons.  A representation that is able to explain many views of the data is more likely to capture meaningful variation than a representation that is a good fit for only one of the views.  They are also attractive for the theoretical reasons. For example, \cite{ak} show that certain classes of latent variable models, such as Hidden Markov Models, Gaussian Mixture Models, and Latent Dirichlet Allocation models, can be optimally learned with multiview spectral techniques.  Representations learned from many views will generalize better than one, since the learned representations are forced to accurately capture variation in all views at the same time \citep{sridharan} -- each view acts as a regularizer, constraining the possible representations that can be learned. These methods are often based on canonical correlation analysis~(CCA), a classical statisical technique proposed by~\cite{cca}.

In spite of encouraging theoretical guarantees, multiview learning techniques cannot freely model nonlinear relationships between arbitrarily many views. Either they are able to model variation across many views, but can only learn linear mappings to the shared space \citep{GCCA1}, or they simply cannot be applied to data with more than two views using existing techniques based on kernel CCA~\citep{hardoon2004canonical} and deep CCA~\citep{originalDCCA}.

Here we present Deep Generalized Canonical Correlation Analysis (DGCCA).  Unlike previous correlation-based multiview techniques, DGCCA learns a shared representation from data with arbitrarily many views and simultaneously learns nonlinear mappings from each view to this shared space.  The only (mild) constraint is that these nonlinear mappings from views to shared space must be differentiable.  Our main methodological contribution is the  derivation of the gradient update for the Generalized Canonical Correlation Analysis (GCCA) objective \citep{GCCA1}.  As a practical contribution, we have also released an implementation of DGCCA\footnote{See \url{https://bitbucket.org/adrianbenton/dgcca-py3} for implementation of DGCCA along with data from the synthetic experiments.}.

We also evaluate DGCCA-learned representations on two distinct datasets and three downstream tasks: phonetic transcription from aligned speech and articulatory data, and Twitter hashtag and friend recommendation from six text and network feature views.  We find that downstream performance of DGCCA representations is ultimately task-dependent.  However, we find clear gains in performance from DGCCA for tasks previously shown to benefit from representation learning on more than two views, with up to 4\% improvement in heldout accuracy for phonetic transcription.

The paper is organized as follows.  We review prior work in Section~\ref{Prior Work}. In Section~\ref{sec:dgcca} we describe DGCCA. Empirical results on a synthetic dataset, and three downstream tasks are presented in Section~\ref{sec:results}. In Section~\ref{sec:other}, we describe the differences between DGCCA and other non-CCA-based multiview learning work and conclude with future directions in Section~\ref{sec:conclusions}.

%% file: dgcca.tex
\section{Deep Generalized Canonical Correlation Analysis (DGCCA)}
\label{sec:dgcca}
In this section, we present deep GCCA (DGCCA): a multiview representation learning technique that benefits from the expressive power of deep neural networks and can also leverage statistical strength from more than two views in data, unlike Deep CCA which is limited to only two views. More fundamentally, deep CCA and deep GCCA have very different objectives and optimization problems, and it is not immediately clear how to extend deep CCA to more than two views. 

\begin{figure}[h]
\begin{center}
\includegraphics[height=0.18
\textheight]{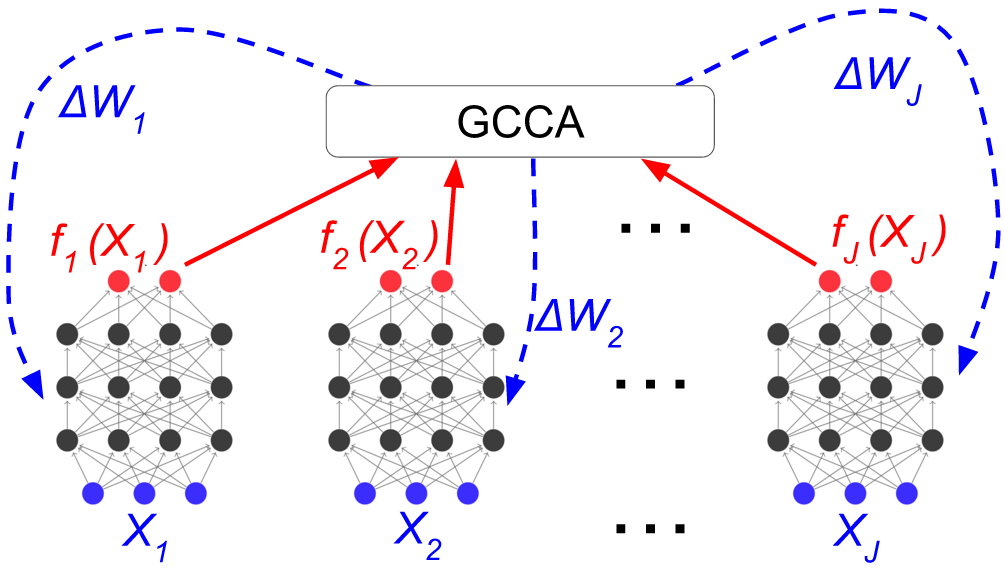}
\end{center}
\caption{A schematic of DGCCA with deep networks for $J$ views.}
\label{NN fig}
 \vspace{-3mm}
\end{figure}

DGCCA learns a nonlinear map for each view in order to maximize the correlation between the learnt representations across views. In training, DGCCA passes the input vectors in each view through multiple layers of nonlinear transformations and backpropagates the gradient of the GCCA objective with respect to network parameters to tune each view's network, as illustrated in Figure \ref{NN fig}.
The objective is to train networks that reduce the GCCA reconstruction error among their outputs. At test time, new data can be projected by feeding them through the learned network for each view.

We now formally define the DGCCA problem. We consider $J$ views in our data, and let $X_j \in \mathbb{R}^{d_j \times N}$ denote the $j^{th}$ input matrix.\footnote{Our notation for this section closely follows that of Andrew et al. \yrcite{originalDCCA}} 
The network for the $j^{th}$ view consists of $K_j$ layers. Assume, for simplicity, that each layer in the $j^{th}$ view network has $c_j$ units with a final (output) layer of size $o_j$. 
\removed{
The output of the $k^{th}$ layer for the $j^{th}$ view is $h^j_k = s(W^j_k h^j_{k-1} + b_k^j), 
$\NoteHK{do we need to explicitly state the bias, or can that be implied} where  $s :  \mathbb{R} \rightarrow  \mathbb{R} $ is a nonlinear activation function, $W^j_{k} \in  \mathbb{R}^{c_k\times c_{k-1}}$ and $b^j_k$ represent the weight matrix and the bias vector, respectively, for the $k^{th}$ layer of the $j^{th}$ view. 
}
The output of the $k^{th}$ layer for the $j^{th}$ view is $h^j_k = s(W^j_k h^j_{k-1})$, where  $s :  \mathbb{R} \rightarrow  \mathbb{R} $ is a nonlinear activation function and $W^j_{k} \in  \mathbb{R}^{c_k\times c_{k-1}}$ is the weight matrix for the $k^{th}$ layer of the $j^{th}$ view network. We denote the output of the final layer as $f_j(X_j)$. 

DGCCA can be expressed as the following optimization problem: find weight matrices $W^j = \{W^j_{1}, \ldots, W^j_{K_j}\}$ defining the functions $f_j$, and linear transformations $U_j$ (of the output of the $j^{th}$ network), for $j=1,\ldots,J$, that 
\vspace*{-10pt}
\begin{align}
\label{eq:dgcca}
\begin{split}
 	\minimize_{U_j \in \mathbb{R}^{o_j \times r}, G \in \mathbb{R}^{r
 	\times N}} &\sum_{j=1}^J
    \|G - U_j^\top f_j(X_j)\|_F^2 
    \\ \text{subject to} \qquad &GG^\top = I_r
\end{split},
\end{align}
where $G \in \mathbb{R}^{r\times N}$  is the shared representation we are interested in learning. 

\vspace*{-5pt}
\paragraph{Optimization:} We solve the DGCCA optimization problem using stochastic gradient descent (SGD) with mini-batches. In particular, we estimate the gradient of the DGCCA objective in Problem~\ref{eq:dgcca} on a mini-batch of samples that is mapped through the network and use back-propagation to update the weight  matrices, $W^j$'s. However, note that the DGCCA optimization problem is a constrained optimization problem. It is not immediately clear how to perform projected gradient descent with back-propagation. Instead, we characterize the objective function of the GCCA problem at an optimum, and compute its gradient with respect to the inputs to GCCA, i.e. with respect to the network outputs. These gradients are then back-propagated through the network to update $W^j$'s.

Although the relationship between DGCCA and GCCA is analogous to the relationship between DCCA and CCA, derivation of the GCCA objective gradient with respect to the network output layers is non-trivial. The main difficulty stems from the fact that there is no natural extension of the correlation objective to more than two random variables. Instead, we consider correlations between every pair of views, stack them in a $J \times J$ matrix and maximize a certain matrix norm for that matrix. For GCCA, this suggests an optimization problem that maximizes the sum of correlations between a shared representation and each view. Since the objective as well as the constraints of the generalized CCA problem are very different from that of the CCA problem, it is not immediately obvious how to extend Deep CCA to Deep GCCA.

Next, we show a sketch of the gradient derivation, the full derivation is given in appendix \ref{sect:deriv}. It is straightforward to show that the solution to the GCCA problem is given by solving an eigenvalue problem. In particular, define $C_{jj} = f(X_j)f(X_j)^\top \in \mathbb{R}^{o_j \times o_j}$, to be the scaled empirical covariance matrix of the $j^{th}$ network output, and $P_j = f(X_j)^\top C_{jj}^{-1} f(X_j) \in \mathbb{R}^{N \times N}$ be the corresponding projection matrix that whitens the data; note that $P_j$ is symmetric and idempotent.  We define $M = \sum_{j=1}^J P_j$. Since each $P_j$ is positive semi-definite, so is $M$. Then, it is easy to check that the rows of $G$ are the top $r$ (orthonormal) eigenvectors of $M$, and $U_j = C_{jj}^{-1}f(X_j)G^\top$. Thus, at the minimum of the objective, we can rewrite the reconstruction error as follows: 
\begin{align*}
\sum_{j=1}^J \|G - U_j^\top f_j(X_j)\|_F^2 &= 
\sum_{j=1}^J \|G - G f_j(X_j)^\top C_{jj}^{-1} f_j(X_j)\|_F^2 
= rJ - \tr(GMG^\top)
\end{align*}

\removed{
In order to minimize the objective and update the weights, $W^j$, we perform backpropagation in each network. In order to do so, we need to take the derivative of the objective in equation \NoteHK{refer to the objective}.
This is, then, used to update the weights, $W^j$ through backpropagation in each network. We show a sketch of the derivation here, the full derivation is shown in the Appendix in section \ref{sect:deriv}. 
}
\removed{
It can be shown that the solution is found by solving a certain
eigenvalue problem. In particular, define $C_{jj} = f(X_j)f(X_j)^\top \in
\mathbb{R}^{d_j \times d_j}$, $P_j = f(X_j)^\top C_{jj}^{-1} X_j$
(note that $P_j$ is symmetric and idempotent), and
$M = \sum_{j=1}^J P_j$ (since each $P_j$ is psd, so is $M$).
Then the rows of $G$ are the top $r$ (orthonormal) eigenvectors of $M$,
and $U_j = C_{jj}^{-1}f(X_j)G^\top$. Thus, at the minima of the objective,
we can rewrite the reconstruction error as follows:
\begin{align*}
\sum_{j=1}^J \|G - U_j^\top f_j(X_j)\|_F^2 &= 
\sum_{j=1}^J \|G - GX_j^\top C_{jj}^{-1}X_j\|_F^2 \\
&= Jr - \tr(GMG^\top)
\end{align*}
}

Minimizing the GCCA objective (w.r.t. the weights of the
neural networks) means maximizing $\tr(GMG^\top)$, which is the sum of eigenvalues $L=\sum_{i=1}^r \lambda_i(M)$.
Taking the derivative of $L$ with respect to each output layer $f_j(X_j)$ we have:
\[
\frac{\partial L}{\partial f_j(X_j)}
= 2U_jG - 2U_jU_j^\top f_j(X_j)
\]
Thus, the gradient is the difference between the
$r$-dimensional auxiliary representation $G$ embedded into
the subspace spanned by the columns of $U_j$ (the first term) 
and the projection of
the actual data in $f_j(X_j)$ onto the said subspace (the second term).
Intuitively, if the auxiliary representation $G$ is far away from
the view-specific representation $U_j^\top f_j(X_j)$, then the
network weights should receive a large update.  Computing the gradient
descent update has time complexity $O(J N r d)$, where
$d = max (d_1, d_2, \ldots , d_J )$ is the largest dimensionality of the input views.